%% file: root.tex
\documentclass[letterpaper, 10 pt, conference]{ieeeconf} 

\input{macros}

\begin{document}

\title{DeepReach: A Deep Learning Approach to High-Dimensional Reachability}
\author{Somil Bansal and Claire J. Tomlin}
\maketitle

\begin{abstract}
Hamilton-Jacobi (HJ) reachability analysis is an important formal verification method for guaranteeing performance and safety properties of dynamical control systems. 
Its advantages include compatibility with general nonlinear system dynamics, formal treatment of bounded disturbances, and the ability to deal with state and input constraints. 
However, it involves solving a PDE, whose computational and memory complexity scales exponentially with respect to the number of state variables, limiting its direct use to small-scale systems.
We propose \metName{}, a method that leverages new developments in sinusoidal networks to develop a neural PDE solver for high-dimensional reachability problems. 
The computational requirements of \metName{} do not scale directly with the state dimension, but rather with the complexity of the underlying reachable tube. 
\metName{} achieves comparable results to the state-of-the-art reachability methods, 
does not require any explicit supervision for the PDE solution, 
can easily handle external disturbances, adversarial inputs, and system constraints, 
and also provides a safety controller for the system.
We demonstrate \metName{} on a 9D multi-vehicle collision problem, and a 10D narrow passage problem, motivated by autonomous driving applications.
\end{abstract}

\IEEEpeerreviewmaketitle

\input{sections/intro}
\input{sections/related_work}
\input{sections/problem_setup}
\input{sections/background}
\input{sections/method}
\input{sections/results}
\input{sections/conclusion}

\appendices

\printbibliography

\end{document}

%% file: macros.tex
\usepackage{times}
\usepackage{color, colortbl, xcolor}
\input{notation}

\usepackage{multicol}
\usepackage[bookmarks=true]{hyperref}

\usepackage{amsfonts}
\usepackage{algorithmic}
\usepackage[ruled,vlined,titlenotnumbered]{algorithm2e} 
\usepackage{amsmath}
\usepackage{dsfont}
\usepackage{subfigure}
\usepackage{multicol}
\usepackage[pdftex]{graphicx}   
\usepackage{graphicx}
\usepackage{bbm}
\usepackage{mathtools}
\usepackage{dsfont}
\usepackage{accents}

\newcommand{\ubar}[1]{\underaccent{\bar}{#1}}

\graphicspath{{./figs/}}    
\DeclareGraphicsExtensions{.pdf,.png}  

\usepackage[textfont=md,font=footnotesize]{caption}

\usepackage[left=55pt, right=55pt, top=55pt, bottom=58pt]{geometry} 

\IEEEoverridecommandlockouts
\usepackage{balance}

\usepackage[%
    style=numeric-comp,
    sorting=none,
    backend=biber,
    sortcites=true,
    doi=false,
    firstinits=true,
    hyperref,
    isbn=false,
    eprint=false,
    maxcitenames=3, 
    minbibnames=3, 
    maxbibnames=4, 
    block=none]
    {biblatex}
    
    \renewbibmacro{in:}{}
    \AtEveryBibitem{%
  	\clearlist{language}%
  	\clearfield{pages}%
	}
\bibliography{references}
\usepackage{lipsum}

\newcommand\blfootnote[1]{%
  \begingroup
  \renewcommand\thefootnote{}\footnote{#1}%
  \addtocounter{footnote}{-1}%
  \endgroup
}

\definecolor{planning_color}{RGB}{69, 174, 254}    
\definecolor{prediction_color}{RGB}{255, 116, 190}
    
\newcommand{\example}[1]%
{
\textbf{Running example:}
\textit{#1}
}



%% file: notation.tex
\newcommand{\tvar}{t}
\newcommand{\tdummy}{\tau}
\newcommand{\R}{\mathbb{R}}
\newcommand{\ctrl}{u}
\newcommand{\dstb}{d}
\newcommand{\cfunc}{u(\cdot)}
\newcommand{\dfunc}{d(\cdot)}
\newcommand{\cset}{\mathcal{U}}

\newcommand{\dset}{\mathcal{D}}

\newcommand{\state}{x}
\newcommand{\traj}{\xi} 

\newcommand{\dyn}{f} 
\newcommand{\targetfunc}{l}
\newcommand{\targetset}{\mathcal{L}}
\newcommand{\obsfunc}{g}
\newcommand{\obsset}{\mathcal{G}}
\newcommand{\costfunctional}{J}

\newcommand{\vfunc}{V}
\newcommand{\vset}{\mathcal{V}}

\newcommand{\trajstandard}{\traj_{\state,\tvar}^{\ctrl,\dstb}}

\newcommand{\param}{\theta}
\newcommand{\nnvfunc}{\vfunc_{\param}}
\newcommand{\loss}{h}

\newcommand{\veh}{Q}

\newcommand{\horizon}{T}

\newcommand{\metName}{\text{DeepReach}}

\newtheorem{remark}{Remark}

%% file: sections/intro.tex
\section{Introduction}
As autonomous systems are integrated in our day-to-day lives, ensuring provable safety guarantees and controllers for these systems is vital.
Hamilton-Jacobi (HJ) reachability analysis is a verification method for autonomous systems that computes both the safe configurations and the corresponding safe controller for the system. 
In reachability analysis, one computes the \textit{Backward Reachable Tube (BRT)} of a dynamical system.
\blfootnote{$^{1}$All authors are with the Department of Electrical Engineering and Computer Sciences, University of California, Berkeley: \{somil, tomlin\}@berkeley.edu.}
\blfootnote{$^{2}$This research is supported by the DARPA Assured Autonomy program under agreement number FA8750-18-C-0101.}
This is the set of states such that the trajectories that start from this
set will eventually reach some given target set despite the worst case disturbance (or an exogenous, adversarial input more generally).
As an example, for an aerial vehicle, the disturbance could be wind or another adversarial aircraft flying nearby, and the target set could be the destination of the vehicle. 
The BRT provides both the set of states from which the aerial vehicle can safely reach its destination and a robust controller for the vehicle.
Conversely, if the target set consists of those states that are known to be unsafe, the BRT represent the states from which the system will end up in the target set for some disturbance, despite the best possible control efforts.
Thus, the BRT contains states which are potentially unsafe and should therefore be avoided.
The converse of BRT in this case provides the set of safe states for the system and the corresponding safety controller.
If the system additionally needs to satisfy state dependent constraints at all times, then it is called a \textit{Backward Reach-Avoid Tube (BRAT)} instead of BRT.

Traditionally, BRT computations in HJ reachability are formulated as a zero-sum dynamic game between the control and the disturbance whose value function can be used to synthesize the BRT and the safety controller.
This involves solving a Hamilton-Jacobi (HJ) partial differential equation (PDE) on a grid representing a discretization of the state space.
Unfortunately, this process becomes computationally intensive for large systems, resulting in an exponential scaling of memory and computational complexity with respect to system dimensionality; this is often referred to as the ``\textit{curse of dimensionality}.''
Several methods have been proposed to overcome this challenge by trading off between the class of dynamics they can handle, the approximation quality of the BRT, and the required computation. 
Additionally, very few of these approaches can handle adversarial inputs; yet, considering adversarial inputs (e.g., actions of other agents or the model and the environment uncertainty) and state constraints is critical in analysis 
and synthesis of safe controllers.

In this work, we propose \metName{}, a deep learning-based approach to approximately solve high-dimensional reachability problems. 
What sets \metName{} apart is its ability to compute BRTs (and BRATs) as well as the corresponding safety controller for \textit{general} nonlinear dynamical systems in the presence of \textit{disturbances} and \textit{state and input constraints}. 
\metName{} is rooted in HJ reachability analysis; however, instead of solving the value function PDE over a grid, \metName{} draws inspiration from the recent progress in solving PDEs using deep learning, and represents the value function as a deep neural network (DNN) to learn a parameterized approximation of the value function.
Thus, the computation and memory requirements for obtaining the value function do not scale with the grid resolution, but rather the complexity of the value function.
    
To overcome the challenges of obtaining the supervision data for learning the value function in high-dimensional systems, we use a self-supervision method for training the DNN. 
The key insight behind our method is that if we use a function approximator whose gradients are well behaved, then the PDE itself can be used to self-supervise the learning of the value function.
However, the widely popular ReLU-based architectures, while promising, lack the capacity to effectively represent the gradients of a target signal. 
This is partly due to the fact that ReLU networks are piecewise linear, their first derivative is piecewise constant, and therefore struggle when encoding a value function with fine details.
To overcome this challenge, we use periodic, sinusoidal activation functions, that have recently shown a lot of promise in representing not only the underlying signal well, but also its derivatives \cite{sitzmann2020implicit}. 
This has two advantages: first, it enables a self-supervision method for computing the value function, allowing us to scale the proposed approach to general dynamical systems.
Second, the safety controller depends on the gradients of the value function and has traditionally been hard to compute using approximate methods.
With \metName{}, we can also compute the optimal safety control which is important in a variety of applications.
We demonstrate the capabilities of \metName{} in computing BRTs and the safety controller for a variety of high-dimensional reachability problems that arise in multi-vehicle collision mitigation and autonomous driving applications.

%% file: sections/related_work.tex
\section{Related Work}

\noindent \textbf{Reachability analysis for dynamical systems.} Methods that compute BRTs accurately up to numerical precision include level set methods \cite{mitchell2005time, lygeros2004reachability, Margellos11, fisac2015reach, bansal2017hamilton}.
Level set methods are rooted in HJ reachability and compute BRTs by solving the value function PDE on a grid representing a discretization of the state space \cite{mitchell2004toolbox}.
These methods can deal with nonlinear dynamics, disturbances, and state constraints; however, their exponentially scaling computational complexity limits their direct use to upto 5D systems \cite{bansal2017hamilton}. 
Recently, there have been attempts to scale these methods to 7D systems by using a very coarse discretization; however, the computed value function quickly accumulates numerical errors \cite{leung2020infusing}. 

Several approximate methods overcome these challenges by making tradeoffs in the class of dynamics they can consider; whether they can handle control, disturbances, and/or constraints; and whether they compute an exact or an approximate BRT.
For example, the methods presented in \cite{Greenstreet1998, Frehse2011, Kurzhanski00, Kurzhanski02, Maidens13, girard2005reachability, althoff2010computing} have had success in analyzing high-dimensional affine systems using sets of pre-specified shapes, such as polytopes, hyperplanes, ellipsoids, and zonotopes.
However, these methods often rely on linearization for nonlinear dynamics \cite{schurmann2017guaranteeing} which might lead to an overly conservative BRT, especially in the presence of disturbances and constraints.
Other methods rely on specific dynamics, including that use sum-of-squares optimization \cite{Majumdar13, majumdar2017funnel} and parallelotopes \cite{Dreossi16} for computing BRTs for polynomial dynamics \cite{henrion2014convex}; leverage monotonicity of dynamics \cite{coogan2015efficient}; use decoupled or reduced order dynamics to decompose the BRT computation \cite{chen2018decomposition, holmes2020reachable, liu2020leveraging}; leverage the Hopf-Lax formula to compute BRTs for convex dynamics \cite{chow2017algorithm}; or deal with affine dynamics \cite{bak2019numerical, Nilsson2016}.
Control Lyapunov (and Barrier) functions are also used to design safety controller \cite{ames2016control, ames2019control}; however, finding feasible Lyapunov functions can be challenging, especially in the presence of control or disturbance bounds.

Sampling and learning-based methods have also been explored for computing reachable tubes \cite{liebenwein2018sampling}, such as the methods that use scenario optimization for computing BRTs \cite{sutter2017data, devonport2020estimating, ghosh2019new} or leverage adversarial sampling to compute a high-confidence convex hull of BRTs \cite{lew2020sampling}.
Other methods use Support Vector Machines \cite{allen2014machine} and DNNs \cite{rubies2019classification, fisac2019bridging, darbon2020overcoming, niarchos2006neural, djeridane2006neural} to approximate the value function and BRTs. 
These methods can leverage data to compute approximate BRTs for nonlinear systems; however, they often struggle in the presence of adversarial disturbances and state constraints.
Other methods require explicit supervision for learning which is hard to generate for general high-dimensional systems.
In contrast, \metName{} does not rely on explicit supervision, and can compute BRTs for nonlinear dynamics in the presence of disturbances.
\metName{} can also be interpreted as an approximate dynamic programming method \cite{bertsekas1995dynamic} where a DNN is used to \textit{learn} the basis features for the value function. 
\\

\noindent \textbf{Neural PDE solvers.} DNNs have long been investigated in the context of solving differential equations \cite{lee1990neural}.
Early work on this topic involved simple neural network models, consisting of MLPs or radial basis function networks with few hidden layers and hyperbolic tangent or sigmoid nonlinearities \cite{lagaris1998artificial, he2000multilayer, mai2003approximation}. 
Modern approaches to these techniques leverage recent advances in deep learning to solve more sophisticated equations with higher dimensionality and more constraints \cite{sirignano2018dgm, raissi2019physics, berg2018unified, han2018solving, chen2018neural}. 
However, as demonstrated recently, the commonly used MLPs with smooth, non-periodic activation functions fail to accurately model derivatives even with dense supervision \cite{sitzmann2020implicit}.
Moreover, the authors demonstrate that periodic activation functions can overcome this challenge.
Our work builds upon this insight to solve high-dimensional PDEs that arise in the reachability problems, without requiring any explicit supervision for the value function.

%% file: sections/problem_setup.tex
\section{Problem Setup} \label{sec:problem_setup}
Consider an autonomous agent in an environment in the presence of external disturbance.
We model the agent as a dynamical system with state $\state \in \R^n$, control $\ctrl$, and disturbance $\dstb$. The state evolves according to the dynamics:
\begin{equation}
\label{eq:dyn}
\begin{aligned}
\dot{\state} = \dyn(\state, \ctrl, \dstb), \quad \ctrl \in \cset, \dstb \in \dset.
\end{aligned}
\end{equation}
The disturbance might be an actual exogenous input or represent the model and environment uncertainty that we want to safeguard against.
Let $\trajstandard(\tdummy)$ denote the state achieved at time $\tdummy$ by starting at initial state $\state$ and initial time $\tvar$, and applying input functions $\cfunc$ and $\dfunc$ over $[\tvar,\tau]$\footnote{We assume that the control and disturbance inputs are drawn from compact sets ($\cset$, $\dset$) and the flow field $\dyn: \mathbb{R}^{n}\times\cset\times\dset \rightarrow \mathbb{R}^{n}$ satisfies the standard assumptions of existence and uniqueness of the state trajectories $\traj(\tdummy; \state, \tvar, \cfunc, \dfunc)$ (or $\trajstandard(\tdummy)$ for compactness) \cite{EarlA.Coddington1955}.}.  
The environment contains a target set $\targetset$ that is important to the agent: it can be either a set of goal states, or a set of unsafe states.
In this work, we are interested in computing the Backward Reachable Tube (BRT) and the Backward Reach-Avoid Tube (BRAT) of $\targetset$.

\textbf{Backward Reachable Tube (BRT).} 
BRT is the set of initial states for which the agent acting optimally and under worst-case disturbances, will \textit{eventually} reach the target set $\targetset$ within the time horizon $[\tvar, \horizon]$:
\begin{equation}
\label{eqn:BRT}
\vset(\tvar) = \{\state: \forall \ctrl(\cdot), \exists \dstb(\cdot), \exists \tdummy \in [\tvar, \horizon], \trajstandard(\tdummy) \in \targetset\}
\end{equation}
If the target set consists of those states that are known to be unsafe, then the BRT contains states which are potentially unsafe and should therefore be avoided.
For goal states, BRT can be similarly defined with the roles of control and disturbance switched.

\textbf{Backward Reach-Avoid Tube (BRAT).} When the target set represents the goal states, BRAT is the set of initial states of the agent from which it can eventually reach the target set, despite the worst-case disturbance and \textit{while avoiding} some unsafe set of states $\obsset$ at all times.
Robot trajectory planning problems fall under this category, where $\targetset$ is the goal configuration of the robot and $\obsset$ represent potential obstacles in the environment (more generally, the set of states where system constraints are violated).
Mathematically,
\begin{align}
\label{eqn:BRAT}
\vset(\tvar) = \{\state: \forall \dstb(\cdot), \exists \ctrl(\cdot), & \forall s \in [\tvar, \horizon], \trajstandard(s) \notin \obsset, \nonumber \\
& \exists \tdummy \in [\tvar, \horizon], \trajstandard(\tdummy) \in \targetset\}.
\end{align}
When no constraints are present in the system, the set in \eqref{eqn:BRAT} reduces to a BRT.

\textit{\textbf{Running example (Air3D).}} We now introduce a collision avoidance problem between two identical vehicles that we will use to illustrate the key ideas. 
The (relative) dynamics between the two vehicles (called evader and pursuer) are:
\begin{equation}
\label{eq:dyn_air3D}
\begin{aligned}
\dot{\state}_1 &=  -v_e + v_p cos \state_3 + \omega_e \state_2\\
\dot{\state}_2 &= v_p sin \state_3 - \omega_e \state_1\\
\dot{\state}_3 &=  \omega_p - \omega_e,
\end{aligned}
\end{equation}
where $\state_1$, $\state_2$ represent the relative position between the vehicles and $\state_3$ represents the relative heading. 
$v_e$ and $v_p$ are the linear velocities of the evader and pursuer respectively. 
These velocities are constant and equal. 
$\omega_e$ and $\omega_p$ are the angular velocities of the vehicles, and are the input and disturbance in the system respectively.
Both $\omega_e$ and $\omega_p$ are bounded to the interval $[-\bar\omega, \bar\omega]$.
We are interested in computing the BRT of the target set
\begin{equation}
\label{eq:lx_air3D}
\begin{aligned}
\targetset = \{\state: \|(\state_1, \state_2)\| \leq \beta\}.
\end{aligned}
\end{equation}
Intuitively, the target set represents the set of all relative states where the evader and the pursuer are in close proximity (i.e., the collision set).  
The BRT is the set of states from which the pursuer can drive the system trajectory into the collision set, despite the best efforts of the evader to avoid a collision.

Even though only three-dimensional, this problem is a popular benchmark in the reachability literature because it features one of the very few nonlinear dynamical systems for which the BRT can be computed (almost) analytically, against which the proposed solution can be compared \cite{mitchell2020robust}. 
Moreover, the BRT is known to be nonconvex with a surface that is non-differentiable in places.
Moreover, very few reachability algorithms can tackle both a control and an adversarial input.
We will present higher dimensional examples in Sec. \ref{sec:examples}.

%% file: sections/background.tex
\section{Background: Hamilton-Jacobi Reachability} \label{sec:hj_reachability}
One way to compute BRT/BRAT is through Hamilton-Jacobi (HJ) reachability.
In HJ reachability, the computation of BRT (or BRAT) is formulated as a zero-sum game between the control and the disturbance.
This zero-sum game is a robust optimal control problem which can be solved using the principle of dynamic programming. 
The BRT can then be recovered from the obtained solution. 

First, a target function $\targetfunc(\state)$ is defined whose sub-zero level set is the target set $\targetset$, i.e. $\targetset = \{\state : \targetfunc(\state) \leq 0\}$.  Typically,  $\targetfunc(\state)$ is defined as a signed distance function to $\targetset$.
The BRT seeks to find all states that could enter $\targetset$ at any point in the time horizon, and therefore become unsafe. 
This is computed by finding the minimum distance to $\targetset$ over time:
\begin{equation}
    \label{eq:costfunctional}
    \costfunctional(\state,\tvar,\cfunc,\dfunc) = \min_{\tdummy \in [\tvar, \horizon]} \targetfunc(\trajstandard(\tdummy)).
\end{equation}
Our goal is to capture this minimum distance for \textit{optimal trajectories} of the system.  
Thus, we compute the optimal control that maximizes this distance (and drives the system away from the unsafe target set) and the worst-case disturbance signal that minimizes the distance. 
The value function corresponding to this robust optimal control problem is:
\begin{equation}
    \label{eq:valuefunc}
    \vfunc(\state,\tvar) = \inf_{\dfunc} \sup_{\cfunc} \Big\{\costfunctional\Big(\state,\tvar,\cfunc, \dfunc\Big)\Big\}.
\end{equation}
The value function in \eqref{eq:valuefunc} can be computed using dynamic programming, which results in the following final value Hamilton-Jacobi-Isaacs Variational Inequality (HJI VI):
\begin{equation}
\begin{aligned}
    \label{eq:HJIVI}
    \min\Big\{D_\tvar \vfunc(\state,\tvar)+ &H(\state,\tvar), \targetfunc(\state)-\vfunc(\state,\tvar)\Big\} = 0, \\
    &\vfunc(\state,\horizon) = \targetfunc(\state).
    \end{aligned}
\end{equation}
$D_\tvar$ and $\nabla$ represent the time and spatial gradients of the value function. 
$H$ is the Hamiltonian, which optimizes over the inner product between the spatial gradients of the value function and the flow field of the dynamics to compute the optimal control and disturbance inputs:
\begin{equation}
    \label{eq:ham}
    \begin{aligned}
    H(\state,\tvar) = \max_\ctrl \min_\dstb & \langle \nabla \vfunc(\state,\tvar), \dyn(\state,\ctrl,\dstb)\rangle.
        \end{aligned}
\end{equation}
Intuitively, \eqref{eq:HJIVI} can be thought of as a continuous time analogue of Bellman equation in continuous time and in the presence of disturbances. 
The term $\targetfunc(\state)-\vfunc(\state,\tvar)$ in \eqref{eq:HJIVI} restricts the system trajectories to enter and then leave the target set, effectively enforcing that all trajectories that achieve negative distance at any time will continue to have negative distance for the rest of the time horizon. 
Once the value function is obtained, the BRT is given as the sub-zero level set of the value function
\begin{equation}
    \label{eq:BRT_from_valfunc}
    \vset(\tvar) = \{\state: \vfunc(\state,\tvar) \leq 0 \}.
\end{equation}
The corresponding optimal safety control can be derived as
\begin{equation}
    \label{eq:opt_ctrl}
    \begin{aligned}
    u^*(\state,\tvar) = \arg\max_\ctrl \min_\dstb & \langle \nabla \vfunc(\state,\tvar), \dyn(\state,\ctrl,\dstb)\rangle.
        \end{aligned}
\end{equation}
In fact, the system can apply any control while still maintaining safety, as long as starts outside the BRT and applies the safety control in \eqref{eq:opt_ctrl} at the BRT boundary.
The optimal adversarial disturbance can be similarly obtained as \eqref{eq:opt_ctrl}.
One can similarly derive a HJI VI for BRAT
\begin{equation}
\begin{aligned}
    \label{eq:HJIVI_BRAT}
    \max\Big\{\min\Big\{&D_\tvar \vfunc(\state,\tvar)+ H(\state,\tvar), \targetfunc(\state)-\vfunc(\state,\tvar)\Big\}, \\
    &\obsfunc(\state)-\vfunc(\state,\tvar)\Big\} = 0, \\
    &\vfunc(\state,\horizon) = \max\big\{\targetfunc(\state), \obsfunc(\state)\big\},
    \end{aligned}
\end{equation}
where $\obsset$ is defined as the super-zero level set of $\obsfunc(\state)$, i.e., $\obsset = \{\state : \obsfunc(\state) > 0\}$.
The outermost max in \eqref{eq:HJIVI_BRAT} ensures that the system trajectories that ever enter $\obsset$ are marked unsafe over the entire horizon.
For more details on the derivation of HJI VI and its variations, please refer to \cite{bansal2017hamilton}.

Typically, solving HJI VI involves computing value function over a grid representing a discretization of state space and time, resulting in an exponential scaling complexity both in computation and memory, limiting its direct use to fairly low-dimensional systems. 
In this work, we will employ machine learning to combat this challenge.
\label{sec:background}

\textit{\textbf{Running example (Air3D).}} $\targetfunc(\state)$ for the running example is  $\targetfunc(\state) := \|(\state_1, \state_2)\| - \beta$.
The Hamiltonian can be computed analytically in this case and it can be shown that:
\begin{equation}
    \label{eq:ham_air3D}
    \begin{aligned}
    H(\state,\tvar) & = p_1(-v_e + v_p cos \state_3) + p_2(v_p sin \state_3) \\
    &- \bar{\omega}\|p_1\state_2 - p_2\state_1 - p_3\| + \bar{\omega}p_3,
    \end{aligned}
\end{equation}
where $p_1$, $p_2$, and $p_3$ are the spatial derivative of the value function with respect to $x_1$, $x_2$, and $x_3$ respectively.

\begin{remark}
In general, the computation of the Hamiltonian in \eqref{eq:ham} requires solving an optimization problem; however, most mechanical systems are control and disturbance affine. For such systems, the Hamiltonian can be computed analytically in terms of value function gradients (e.g., see \eqref{eq:ham_air3D}).
\end{remark}

%% file: sections/method.tex
\section{\metName: Reachability Using Deep Learning}
We propose a deep learning-based approach to solving high-dimensional reachability problems, which we refer to as \textit{\metName}. 
\metName{} leverages a deep neural network (DNN) to represent the value function in \eqref{eq:HJIVI}. 
The input to the DNN is a state vector $\state$ and a time point $\tvar$, and its output is the corresponding value $\vfunc_{\param}(\state,\tvar)$, where $\param$ represent the parameters of the DNN.
The key benefit of representing the value function through a DNN is that DNNs are agnostic to grid resolution, and the memory required generally scales with signal complexity, independent of spatial resolution.
However, learning the parameters $\param$ can be challenging in general, and might require supervision for $\vfunc(\state,\tvar)$ which itself is hard to generate for high-dimensional systems in the first place. 
To overcome this challenge, we propose a self-supervision method to learn the value function implicitly, using HJI VI in \eqref{eq:HJIVI} as the source of supervision. 
Given an input $(x_i, t_i)$, the loss function $\loss(x_i, t_i; \param)$ for training the DNN is given as:
\begin{equation}
\begin{aligned}
    \label{eq:loss_function}
    \loss(x_i, t_i; \param) & = \loss_1(x_i, t_i; \param) + \lambda \loss_2(x_i, t_i; \param), \\
    \loss_1(x_i, t_i; \param) & = \|\nnvfunc(\state_i,t_i) - \targetfunc(\state_i)\| \mathds{1}(t_i = \horizon),\\
    \loss_2(x_i, t_i; \param) & = \|\min\Big\{D_\tvar \nnvfunc(\state_i,\tvar_i)+ H(\state_i,\tvar_i), \\
    &~~~~~~~~~~~~~~~\targetfunc(\state_i)-\nnvfunc(\state_i,\tvar_i)\Big\}\|.
    \end{aligned}
\end{equation}
Intuitively, $\loss_2$ incentivizes the DNN to learn a value function that is \textit{consistent} with the HJI VI without explicitly requiring the supervision for the value function.
$\loss_1$ uses the ground truth value function (available only at the terminal time) to additionally force the DNN to learn the correct value function at the terminal time, thus avoiding learning any degenerate value functions (such as constant value functions which will also satisfy the HJI VI).
$\lambda$ trades-off the relative importance of the two loss functions.
Together, the two loss functions encourage the DNN to learn a value function that satisfies the HJI VI as well as the boundary condition.
The loss function for BRAT can be defined analogously to \eqref{eq:loss_function} by using the HJI VI in \eqref{eq:HJIVI_BRAT}. 

The loss function in \eqref{eq:loss_function} depends on the time and spatial gradients of the value function; thus, the DNN should not only represent the value function well but also its gradients.
The gradients of the value function are also required at the inference time to obtain the optimal safe controller (see Eqn. \eqref{eq:ham}). 
As we demonstrate further in Sec. \ref{sec:examples}, widely popular, ReLU-based DNNs struggle to accurately represent the gradients of the value function, leading to a poor approximation of the value function.
Our key insight is to use a sinusoidal activation function, that has recently shown a lot of promise in representing not only the target signal but also its derivatives \cite{sitzmann2020implicit}, along with the loss function in \eqref{eq:loss_function} to learn a high quality approximation of the value function, without explicitly solving the HJ PDE.

\textbf{Training Procedure.} To facilitate the value function learning using sinusoidal networks, we first pre-train the DNN to learn the correct value function at the terminal time. $\lambda$ is set to zero during this training phase.
We then perform curriculum learning over the time interval -- slowly decrease the value of $\tvar$ linearly as training progresses, starting from $\horizon$. 
This allows the terminal condition to slowly propagate to the earlier time points.
At each training iteration, we sample $N$ states from the state space uniformly randomly and optimize the DNN parameters to minimize the loss function $\loss$, until a desired number of training iterations are completed. 

%% file: sections/results.tex
\section{Case Studies} \label{sec:examples}
\subsection{Running example: Air3D} \label{sec:air3D}
We now apply \metName{} to the two vehicle collision avoidance example discussed in Sec. \ref{sec:problem_setup}. 
We will first evaluate if \metName{} can approximate the value function and BRT well. 
We will additionally compare the learning performance for different activation functions.
We start with some implementation details.

\noindent \textbf{Data.} The dataset is composed of randomly sampled 3D state coordinates, with $N$ = 65k. During training, we scale all states to $[-1, 1]$ and time to $[0, 1]$, which we find improves convergence. The rest of the dynamics parameters are chosen as: $\horizon = 1s, \beta=0.25m, v_e = v_p = 0.75 m/s, \bar{\omega} = 3rad/s$.  

\noindent \textbf{Architecture.} For all experiments, \metName{} (and baselines) uses a 3 layer DNN, with a hidden layer size of 512. 
We use sine non-linearity in \metName{}. We discuss the effect of non-linearity later in this section.

\noindent \textbf{Hyperparameters.} We set $\lambda$ in \eqref{eq:loss_function} to make each component of the loss approximately equal at the beginning of training.
We pre-train the network for 10K iterations and then perform curriculum training for 100k iterations.
We use the Adam optimizer with a learning rate of $10^{-4}$.

\noindent \textbf{Runtime.} Training required approximately 16 hours to fit the value function using PyTorch with a single GPU worker. 
\\

\begin{figure}[ht]
    \centering
    \includegraphics[width=.95 \columnwidth]{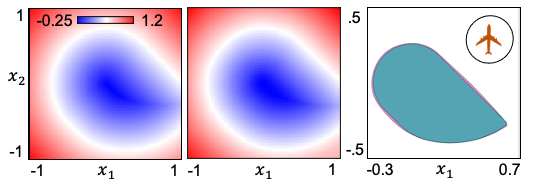}
    \caption{The slices of the value functions and BRTs for $x_3 = \pi/2$. \metName{} recovers a value function (middle) that corresponds closely to the ground truth value function (left), computed using a principled HJI VI solver \cite{mitchell2004toolbox}. 
    The two BRTs also align closely (right) -- they overlap in the green region, the pink region is the error between the two BRTs.}
    \label{fig:air3D_valfn_comparison}
\end{figure}

\metName{} approximates the BRT with a high accuracy, achieving a small MSE of  $1.9 \times 10^{-4}$ with respect to the analytical solution (computed using \cite{mitchell2020robust}).
However, the analytical solution only provides the boundary of the BRT; to analyze the quality of the learned value function, we also solve the HJI VI using a state-of-the-art PDE solver, \textit{Level Set Toolbox (LST)} \cite{mitchell2004toolbox}.  
LST computes the value function over a discrete grid of size $101 \times 101 \times 101$.
A slice of the corresponding value functions for $x_3 = \pi/2$ is shown in Fig. \ref{fig:air3D_valfn_comparison} (left and middle).
\metName{} is able to recover a value function that closely mimics the LST value function and achieves a MSE of $1.01 \times 10^{-4}$.

We also demonstrate the corresponding BRTs in Fig. \ref{fig:air3D_valfn_comparison} (right), overlaid on top of each other.
Even though there is a small error in the boundary of BRT obtained using \metName{}, the two boundaries mimic each other closely. 
We also compute the \textit{BRT volume error} -- the percentage volume of unsafe states that are marked safe (i.e., the set of states that should be in the BRT but are not, and vice-versa) -- as this percentage directly affects the quality of the safety controller.
The BRT volume error for \metName{} is $0.43 \%$, indicating the high quality of the obtained BRT.
\\

\begin{figure}[ht]
    \centering
    \includegraphics[width=.95 \columnwidth]{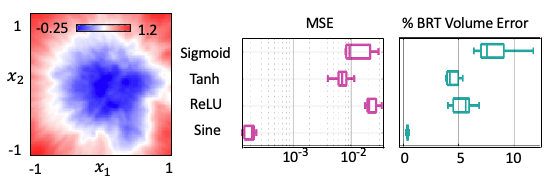}
    \caption{(Left) The slice of the value functions for $x_3 = \pi/2$ corresponding to ReLU activation function. The ReLU architecture fails to learn the correct value function (compare with Fig. \ref{fig:air3D_valfn_comparison} left). 
    This is also evident from high MSE (middle) and \% BRT volume error (right) metrics corresponding to other, non-periodic activation functions.} 
    \label{fig:air3D_nonlinearity_comparison}
\end{figure}

\textbf{Effect of Activation Function.} To analyze the importance of using a periodic activation function in \metName{}, we replaced the sine activation function with ReLU, tanh, and sigmoid functions.
For each activation function, we initialized the learning process with 5 different seeds. 
The corresponding MSE (with respect to the analytical solution) as well as the BRT volume error is reported in Fig. \ref{fig:air3D_nonlinearity_comparison}.
While \metName{} closely matches the analytical solution, the other architectures fail to find the correct solution, with a MSE that is an order of magnitude higher than \metName{}. 

\subsection{Two-Vehicle Collision Avoidance Example} \label{sec:1E1P}
The two vehicle collision avoidance example in Sec. \ref{sec:air3D} is solved in the relative coordinates between the pursuer and the evader; however, the same game can be posed in the joint state space of the two vehicles, leading to a 6D reachability problem. 
This 6D reachability problem is intractable to solve directly through a principled PDE solver and thus cannot be directly compared with \metName{}.
However, we can learn the 6D value function using \metName{} and project the learned value function back to relative coordinates to compare the quality of the obtained solution.
The dynamics of the evader and the pursuer are given as:
\begin{equation}
\label{eq:dyn_1E1P}
\begin{aligned}
\dot{\state}_1 =  v_p cos \state_3\quad \dot{\state}_2 = v_p sin \state_3 \quad \dot{\state}_3 =  \omega_p,\\
\dot{\state}_4 =  v_e cos \state_6\quad \dot{\state}_5 = v_e sin \state_6 \quad \dot{\state}_6 =  \omega_e.
\end{aligned}
\end{equation}
The control of the overall system is $\omega_e$ and the disturbance is $\omega_p$. 
The target set is given by:
\begin{equation}
\label{eq:lx_1E1P}
\begin{aligned}
\targetset = \{\state: \|(\state_1, \state_2) - (\state_4, \state_5)\| \leq \beta\}.
\end{aligned}
\end{equation}

The implementation details are same as in Sec. \ref{sec:air3D}.
We pre-train the network for 10k iterations, followed by 100k iterations of curriculum training, requiring approximately 17 hours to fit the value function using \metName{}. 
We project the obtained value function to relative coordinates and compare it with the analytical solution.
Even though the underlying system has twice as much dimensions now, the performance is on par with the learned solution for the 3D system (MSE of $2.1 \times 10^{-4}$).
It is interesting to note that in addition to maintaining the accuracy of the learned solution, the training time as well as memory requirements for approximating the value function do not increase much from the 3D system to the 6D system for \metName{}.
This indicates that unlike typical HJI VI solvers the proposed approach is agnostic to spatial resolution, and rather the complexity scales with the signal complexity (which is same for the 3D and the 6D system in this case).

\subsection{Three-Vehicle Collision Avoidance Example} \label{sec:2E1P}
We now extend \metName{} to a three-vehicle collision avoidance problem, consisting of two evaders ($e_1$ and $e_2$) and one pursuer ($p$). 
The dynamics of the evaders and the purser are given as in \eqref{eq:dyn_1E1P}, leading to a 9D system in the joint state space of the evaders and the pursuer.
The pursuer here represents the adversarial agent that tries to steer evaders to collide either with the pursuer or with each other; evaders on the other hand try to avoid the collision.
The target set is given as:
\begin{equation*}
\label{eq:lx_2E1P}
\begin{aligned}
\targetset = \{\state: & \|(\state_1, \state_2) - (\state_4, \state_5)\| \leq \beta \vee \|(\state_1, \state_2) - (\state_7, \state_8)\| \\ 
& \leq \beta \vee \|(\state_4, \state_5) - (\state_7, \state_8)\| \leq \beta\},
\end{aligned}
\end{equation*}
where $(\state_1, \state_2, \state_3)$,  $(\state_4, \state_5, \state_6)$, and $(\state_7, \state_8, \state_9)$ represent the states of $e_1$, $e_2$, and $p$ respectively. 
The target set can be compactly represented as: $\targetset = \{\state: d(e_1, e_2, p) \leq \beta\}$, where $d(e_1, e_2, p)$ is the minimum pairwise distance between the three vehicles.
The BRT corresponding to the above target set thus represent the set of unsafe states for the evaders.
The implicit target function $\targetfunc(\state)$ is given by
\begin{equation}
\label{eq:lx_func_2E1P}
\begin{aligned}
\targetfunc(\state) = \min\{& \|(\state_1, \state_2) - (\state_4, \state_5)\|, \|(\state_1, \state_2) - (\state_7, \state_8)\|, \\
& \|(\state_4, \state_5) - (\state_7, \state_8)\|\} - \beta,
\end{aligned}
\end{equation}

To avoid the computational challenges associated with the high-dimensional system, the BRT is typically obtained by computing pairwise BRTs between all possible vehicle pairs, each of which is a lower dimensional reachability problem and can be solved in relative coordinates for instance (Sec. \ref{sec:air3D}).
The union of the pairwise BRTs is then used as an approximation for the BRT of the overall system \cite{dhinakaran2017hybrid}; however, this process often leads to an overly optimistic approximation of the unsafe states as this method fails to capture the three way interactions between the vehicles. 
In particular, an escape (safe) strategy for an evader in a pairwise pursuit-evasion game may no longer be safe in the presence of an additional evader, as it might lead to a collision with the other evader.
Thus, the set of unsafe states (i.e., the BRT) is typically larger than the union of the pairwise BRTs.
With \metName{} we can solve the full 9D pursuit-evasion game and can capture these three way interaction states, which has remained a challenge until now.
\begin{figure}[ht]
    \centering
    \includegraphics[width=0.98\columnwidth]{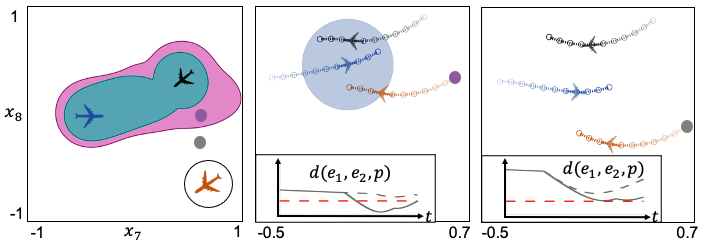}
    \caption{
    The slice of the BRT and example trajectories for the three-vehicle collision avoidance problem. 
    (Left) The pink region represent the set of states that are unsafe because of the three way interaction between vehicles, and were hard to capture previously with low-dimensional approximations (the green region). 
    We mark two starting positions of the pursuer (purple and grey dots) and show the corresponding optimal trajectories for the three vehicles in the middle and the right panel respectively.
    (Middle) When the pursuer starts inside the pink region, it is able to force the two evaders to collide with each other, despite their best efforts to avoid a collision.
    (Right) When the pursuer starts outside the pink region, the evaders are able to barely, but successfully, avoid a collision by applying the safety controller.
    The minimum pairwise distance between the vehicles is shown in inlets.}
    \label{fig:2E1P_brt}
\end{figure}

Given the higher dimensional system, we pretrain the network for 60k iterations, followed by training for 100k iterations, requiring a total of 20 hours to train the network. 
Though, we note that the results to follow are very similar for 30k pretraining iterations. 
The training hyperparamters are same as Sec. \ref{sec:air3D}. 
In Fig. \ref{fig:2E1P_brt} (left) we illustrate a slice of the obtained BRT (the shaded region) for a particular position and orientation of the two evaders (the blue and black aircrafts), and a given heading of the pursuer (the direction of the orange aircraft). 
The green region is the union of the pairwise BRTs, and the pink region is the set of additional states of the pursuer that are unsafe because of the three way interaction between the vehicles. 
\begin{figure*}[ht!]
    \centering
    \includegraphics[width=\textwidth]{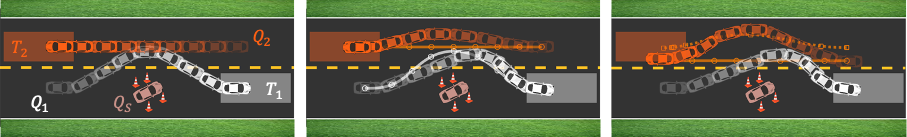}
    \caption{
    The nominal and safe trajectories of the vehicles for the narrow passage problem around a stranded vehicle (earlier timesteps are shown more transparent).
    The white vehicle ($\veh_1$) needs to nudge around the stranded vehicle ($\veh_S$, the purple car) to reach its target ($\mathcal{T}_1$, the white shaded area) while ensuring that it does not collide with the oncoming traffic ($\veh_2$, the orange car). 
    (Left) Following the nominal trajectories leads to a collision.
    (Middle) Safety controller computed using \metName{} ensures that $\veh_1$ undercuts its nominal trajectory and $\veh_2$ swerves around to increase the clearance between the two vehicles and avoid a collision. 
    (Right) The same safety controller automatically figures out that $\veh_2$ needs to swerve more aggressively if it starts closer to the centerline.}
    \label{fig:narrowpassage_scenario1}
\end{figure*}

In Fig. \ref{fig:2E1P_brt} (middle and right), we illustrate the vehicle trajectories when the pursuer starts at the purple and the grey marked positions respectively and when both the evaders and the pursuer use the optimal controller (transparency is higher for the earlier time steps).
In the inlets, we show the minimum across all pairwise distances between the vehicles when the full 9D value function is used (solid gray line) and when the pairwise value function is used (dashed gray line). 
The collision threshold $\beta$ is shown in the dashed red line.
When the pursuer starts inside the pink region, it is able to force the evaders to collide with each other, as indicated by the distance between the vehicles dropping below $\beta$ (middle inlet). 
We show one such collision time in the trajectory plot above -- the shaded blue region represent the set of states around the blue vehicle that are considered to be in collision with the blue vehicle, and the black vehicle is in that region. 
On the other hand, if we simulate the trajectories of the two evaders using the pairwise BRT, no collision occurs between the vehicles (the dashed grey line is above the dashed red line at all times), indicating that these states are unsafe due to the three way interaction between the vehicles.
In contrast, when the pursuer starts just outside the pink region, the evaders has a strategy to avoid the collision despite the best efforts of the pursuer (Fig. \ref{fig:2E1P_brt}, right). 
This is also evident by the minimum distance between the vehicles being just above the dashed red line at all times.
Overall, \metName{} is able to accurately capture the interactions between the vehicles to compute the set of joint unsafe states.

\subsection{Narrow Passage Problem in Autonomous Driving}
We now illustrate how \metName{} can also be used to compute BRATs.
Our problem (depicted in Fig. \ref{fig:narrowpassage_scenario1}) is motivated by autonomous driving application, where the white car ($\veh_1$) encounters a stranded vehicle ($\veh_S$, the purple vehicle) in its lane.
The vehicle thus needs to change its lane (the yellow boundary) to nudge around the stranded car; however, it needs to coordinate this process with the on coming vehicle in the other lane ($\veh_2$, the orange car) in order to avoid a collision with the oncoming traffic or the stranded vehicle. 
We use \metName{} to compute BRAT -- the set of all joint states of the white and the orange car from which they can reach their respective targets ($\mathcal{T}_1$  and $\mathcal{T}_2$ respectively) while avoiding a collision with each other or the stranded vehicle, and without wandering offroad (the grass region). 
More importantly, we are also interested in computing the corresponding safe controllers.

The dynamics of the vehicle $i$, $\veh_i$, are given as:
\begin{equation}
\label{eq:dyn_narrowpassage}
\begin{aligned}
\dot{\state}_{1i} & = {\state}_{4i}~cos(\state_{3i}), \\
\dot{\state}_{2i} & = {\state}_{4i}~sin(\state_{3i}), \\
\dot{\state}_{3i} & = {\state}_{4i}~tan(\state_{5i})/L, \\
\dot{\state}_{4i} & = a_i, \\
\dot{\state}_{5i} & = \psi_i,
\end{aligned}
\end{equation}
where $(\state_1, \state_2)$ represent the position of the vehicle, $\state_3$ is its heading, $\state_4$ is its velocity, and $\state_5$ is its steering angle.
$L$ is the length of the vehicle. 
The control inputs of the vehicle are its acceleration and steering rate, i.e., $u_i = [a_i, \psi_i]$, both of which are bounded within the range $[\ubar{a}, \bar{a}]$ and $[\ubar{\psi}, \bar{\psi}]$ respectively.  
The target function is given as
\begin{equation}
\label{eq:lx_narrowpassage}
\targetfunc(\state) = \max\{d(\veh_1, \mathcal{T}_1), d(\veh_2, \mathcal{T}_2)\},
\end{equation}
where $d(\veh_i, \mathcal{T}_i)$ represent the signed distance between $\veh_i$ and $\mathcal{T}_i$.
The obstacle function $\obsfunc(\state)$ is given by
\begin{equation}
\label{eq:gx_narrowpassage}
\begin{aligned}
\obsfunc(\state) = -\min\{d(\veh_1, \veh_2, \veh_S), d(\veh_1, \mathcal{C}), d(\veh_2, \mathcal{C})\},
\end{aligned}
\end{equation}
where $\mathcal{C}$ represent the road curb boundaries and $d(\veh_1, \veh_2, \veh_S)$ is the minimum pairwise distance between the three vehicles.

We compute the 10D BRAT using \metName{}. 
We pretrain the network for 60k iterations followed by curriculum training over 100k iterations, taking a total of 25 hours to learn the value function.
The rest of the network hyperparameters are same as Sec. \ref{sec:air3D}.

We apply the corresponding safety controller for a variety of scenarios in Fig. \ref{fig:narrowpassage_scenario1}.
The nominal trajectories of the two vehicles are shown in the left panel. 
Following the nominal trajectories without invoking any safety controller leads to a collision between the two vehicles.
In the middle panel, we show the vehicle trajectories in the presence of safety controller -- whenever the joint state of the system reaches on the boundary of BRAT, we override the nominal controller with the safety controller. 
The nominal trajectories are shown in solid white and orange lines respectively.
In the presence of the safety controller, $\veh_2$ swerves towards the road curb to create more room for $\veh_1$ to pass; at the same time, $\veh_1$ takes a tighter turn around the stranded vehicle so as to increase the vertical clearance between $\veh_1$ and $\veh_2$ and avoid a collision. 
Once the two vehicles pass each other, they converge back to their nominal trajectories and reach their respective targets.
This cooperative, collision-avoidance behavior emerge automatically from the safety controller to avoid the collision. 

One of the advantages of reachability is that we don't need to recompute the value function and the safety controller from scratch for every new scenario. 
Since the HJ reachability is based on dynamic programming, it simultaneously provides the safety controller for all possible joint states of the two vehicles. 
For example, in the right panel, we demonstrate how the same value function can be used to compute the safety controller for a different scenario. 
Here, $\veh_1$ commits to follow the same trajectory as in the middle panel and does not use the safety controller.
Moreover, $\veh_2$'s nominal trajectory is now closer to the centerline (the orange solid line). 
Consequently, the safety controller automatically ensures that $\veh_2$ swerves very aggressively towards the road  boundary (compared to its trajectory in the middle panel, shown in the dashed orange line) in order to avoid a collision with $\veh_1$.

%% file: sections/conclusion.tex
\section{Discussion and Future Work}
We propose \metName{}, a method that builds upon recent advances in neural PDE solvers to compute various definitions of reachable tubes as well as a safety controller for dynamical systems.
Unlike traditional approaches, \metName{} does not compute BRTs by solving a PDE explicitly, but rather implicitly by using the PDE as a source of supervision, making it more scalable to higher dimensional systems. 

Even though not exponentially scaling, computing BRTs using \metName{} is still primarily suitable for offline computations.
Moreover, the learning module will inevitably make prediction error, making it challenging to provide any theoretical safety guarantees.
In future, it will be interesting to use \metName{} with online methods that rely on offline, high-dimensional reachability computations to efficiently obtain BRTs in real-time \cite{herbert2017fastrack}, or the methods that ``warm-start'' the BRT computation from a coarse solution to efficiently refine the value function to provide safety guarantees \cite{herbert2019reachability, bajcsy2019safeNavigation}. 

Second, even though we don't expect \metName{} to overcome the curse of dimensionality in general, 
as it is fundamental limitation to any dynamic programming-based method.
However, 
the success of \metName{} on 9D and 10D systems indicates that there might exist a low-dimensional structure to the value function for many common robotic systems that we will exploit further in future for other robotic systems to combat computational complexity associated with BRTs.